\documentclass[10pt, a4paper]{article}

\usepackage[]{lrec2026} 

\usepackage{fancyvrb}
\usepackage{xcolor}
\usepackage{framed}

\definecolor{promptgray}{gray}{0.95}

\newenvironment{promptbox}
  {
   \MakeFramed{\advance\hsize-\width \FrameRestore}
   \scriptsize
   \color{black!90}
  }
  {\endMakeFramed}

\title{Do Language Models Know Theo Has a Wife? \\Investigating the Proviso Problem}

\name{
Tara Azin\textsuperscript{1,$*$,$\dagger$},
Daniel Dumitrescu\textsuperscript{2,$*$},
Diana Inkpen\textsuperscript{2},
Raj Singh\textsuperscript{1}
}

\address{
\textsuperscript{1}Carleton University, Canada \\
\textsuperscript{2}University of Ottawa, Canada \\
$\dagger$Corresponding author: taraazin@cmail.carleton.ca\\
\textsuperscript{$*$}Equal contribution\\
}

\abstract{
We investigate how language models handle the proviso problem, an unresolved issue in pragmatics where presuppositions in conditional sentences diverge between theoretical and human interpretations. We reformulate this phenomenon as a Natural Language Inference task and introduce a diagnostic dataset designed to probe presupposition projection in conditionals. We evaluate RoBERTa, DeBERTa, LLaMA, and Gemma using explainability analyses. The results show that models broadly align with human judgments but rely on shallow pattern matching rather than semantic or pragmatic reasoning. Our work provides the first computational evaluation framework for the proviso problem and highlights the need for diagnostic, multi-method approaches to assess pragmatic competence and context-dependent meaning in language models.
\\ \newline \Keywords{presupposition projection, proviso problem, pragmatic reasoning, language model evaluation, explainability} }

\begin{document}
\pagestyle{plain}
\thispagestyle{plain}
\pagenumbering{arabic}

\maketitleabstract

\section{Introduction}

\label{sec:introduction}

Pragmatics, or the study of context in linguistics, remains an underexplored area in the evaluation of Large Language Models (LLMs). One core pragmatic phenomenon is presupposition, where part of an utterance is assumed to be true and taken as shared knowledge between interlocutors \citep{Strawson-1950, Stalnaker-1973, Stalnaker-1998}. For example, in the sentence \textit{Theo stopped smoking}, the common ground shared by speaker and listener is that Theo used to smoke (the presupposition). Presuppositions are typically signaled by specific lexical items or constructions known as triggers (e.g., stop, again, possessive pronouns).

While straightforward in simple cases, presupposition becomes more complex in embedded structures, such as in conditional sentences. According to formal semantic theories \citep{karttunen-peters-1979-conventional, heim-1992-presupposition, Geurts-1996}, in conditionals of the form $S = \textit{If } A, B_{\textit{p}}$, where $p$ is the direct presupposition of the consequent $B$, the presupposition of the sentence is predicted to project conditionally\footnote{Projection here refers to the persistence of presuppositions under embedding in larger structures, such as negation or conditionals \citep{Karttunen-1973, Heim-1983}.}. For example, in the sentence \textit{If Theo hates sonnets, so does his wife}, the predicted presupposition is \textit{If Theo hates sonnets, then Theo has a wife}.

\begin{figure}[t]
    \centering
    \includegraphics[width=\columnwidth]{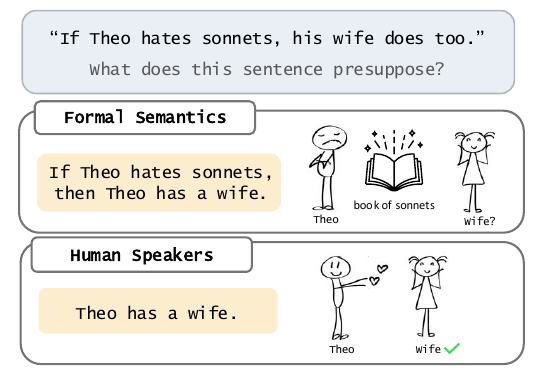}
    \caption{The proviso problem illustrated. Given the sentence \textit{If Theo hates sonnets, so does his wife}, formal semantic theories predict a conditional presupposition, while human speakers typically accommodate an unconditional presupposition. Our work investigates where language models fall in this theory-human divide.}
    \label{fig:stick-figure}
\end{figure}

However, in practice, speakers typically accommodate a stronger, unconditional presupposition: \textit{Theo has a wife}. This phenomenon in pragmatics and formal semantics is called the proviso problem \citep{Geurts-1996}, which remains unresolved \footnote{The problem is often explained by presupposition accommodation, where the listener updates the context to satisfy the sentence’s projected semantic presupposition \citep{Lewis-1979, vonFintel-2008, Singh-2020}.}. This discrepancy between what \textit{should be} presupposed and what speakers \textit{prefer} to presuppose motivates our investigation of how language models handle presuppositions in conditional structures (Figure \ref{fig:stick-figure}).

To the best of our knowledge, our work is the first systematic evaluation of how language models address the proviso problem. We examine whether language models project presuppositions in conditionals as predicted by formal semantic theories or in ways that align with human judgments. To do this, we formulate the proviso problem as a Natural Language Inference (NLI) task. NLI is a well-established framework for modeling inferential relations between a premise and a hypothesis with labels such as Entailment, Neutral, and Contradiction \citep{Bowman-2015,Williams-2018}. In our formulation, the premise takes the form $S = \textit{If } A, B_{\textit{p}}$, where $p$ is the presupposition of $B$, and the hypothesis is p itself. We then assign NLI labels: Entailment (E) when p follows as the presupposition of S, or Neutral (N) when no inferential relationship holds. For example, given the premise \textit{If Theo hates sonnets, so does his wife} and hypothesis \textit{Theo has a wife}, the theory-based label would be N, as the conditional presupposition does not unconditionally entail the hypothesis. In contrast, humans assign E, reflecting the unconditional presupposition that speakers typically accommodate. The contrast between theory-based predictions and human judgments in our evaluation is intended as a methodological design choice for probing model behavior, rather than as a claim that existing theoretical accounts of presupposition projection are inadequate.

For evaluation, we use explainability techniques, such as gradient-based saliency and attention analysis \cite{Sundararajan-2017,Atanasova-2020}, to examine how models come up with their judgments regarding the relation between conditional sentences and their presuppositions. While accuracy metrics show whether a model selects the correct label, they do not make clear whether the decision is driven by sensitivity to presupposition triggers or by reliance on superficial patterns in the data. Our evaluation method addresses this gap by tracking which lexical items and structural features influence model predictions, helping determine whether model behavior aligns with pragmatic reasoning or instead shows reliance on surface-level heuristics.

Our main contributions are as follows:  
\begin{itemize}  
    \item We reformulate the proviso problem as an NLI task to make presupposition projection computationally testable.
    
    \item We create the first dataset for the proviso problem with approximately 8,500 examples featuring structural, semantic, and contextual variations.
    
    \item We evaluate neural language models against theoretical predictions and human judgments using saliency methods to analyze their presuppositional reasoning.
    
\end{itemize}

\section{Related Work}
\label{sec:related-work}

The evaluation of presuppositional reasoning in language models is a relatively recent area of investigation. Early work has examined logical inference in LLMs, including reasoning about conditionals \citep{holliday-etal-2024-conditional} and presupposition judgment \citep{atwell-etal-2025-measuring}, primarily through prompt-based approaches without model fine-tuning. Most prior work on presupposition evaluation has framed the task in the NLI format, creating datasets designed for general inference rather than specifically targeting presuppositional phenomena. Datasets such as IMPPRES \citep{jeretic-etal-2020-natural} and NOPE \citep{Parrish-2021} include some presuppositional content, but they either focus on simple conditional forms or lack coverage of structurally complex cases such as embedded conditionals. 

More recently, CONFER \citep{azin2025confer} introduced a dedicated NLI benchmark for evaluating presuppositional reasoning in conditionals. The dataset includes five types of conditional sentences with human annotations based on the relationship between premises and hypotheses. Their results demonstrate that fine-tuning on this data improves model performance on presuppositional reasoning tasks. However, existing evaluation methods rely primarily on classification accuracy and do not examine whether models process presupposition triggers in ways that align with human reasoning or whether structural variations affect presupposition projection. 

Building on this foundation, our work addresses this methodological gap by combining classification-based evaluation with attribution-driven saliency and attention analyses to examine how models process presuppositions in conditional contexts.

\section{Dataset Construction}
\label{sec:dataset-construction}

We built our dataset on top of CONFER \citeplanguageresource{CONFER}, which contains conditional sentences paired with their presuppositions. From this dataset, we selected 900 sentence pairs. In each pair, the premise follows the form $S = \textit{If } A, B_{\textit{p}}$, where $p$ represents the presupposition of $B$, and the hypothesis restates $p$ itself. Each selected pair is annotated as an entailment relation between the premise and the hypothesis. The conditional sentences contain presupposition triggers \textit{again} and possessive pronouns (e.g., her, him).

In this initial subset, half of the examples show a logical dependency between the antecedent $A$ and the hypothesis $p$. We refer to these as \textit{dependent} cases (DEP), where the content of $A$ semantically relates to or contextually supports $p$. In the other half, $A$ and $p$ are logically unrelated, which we label as \textit{independent} cases (IND), where $A$ describes an unrelated event or circumstance. Distinguishing between DEP and IND examples follows existing dynamic semantics theories, which suggest that the relation between the components of conditionals directly affects whether listeners interpret the presupposition conditionally or unconditionally \citep{Singh2007, vonFintel-2008}. Examples are shown in Table~\ref{tab:dep-ind-examples}. 

We operationalize DEP versus IND as an intra-sentential licensing relation between the antecedent $A$ and the presupposition $p$ of the consequent $B$: DEP if $A$ plausibly supports $p$, IND if it does not. Because our materials are single-sentence pairs, this notion of support is restricted to what is inferable within the sentence.

\begin{table}[ht]
\centering
\small
\begin{tabularx}{\columnwidth}{@{}lX@{}}
\toprule
\multicolumn{2}{c}{\textbf{Dependent case (DEP)}} \\
\midrule
Premise & \textit{If Randolf is a carpenter, he'll use \underline{his} beading tools for designing.} \\
Hypothesis & \textit{Randolf has beading tools.} \\
\midrule
\multicolumn{2}{c}{\textbf{Independent case (IND)}} \\
\midrule
Premise & \textit{If Lisa finishes the meeting early, she'll never drive a sports car \underline{again}.} \\
Hypothesis & \textit{Lisa has driven a sports car before.} \\
\bottomrule
\end{tabularx}
\caption{\label{tab:dep-ind-examples}Examples of DEP and IND cases. In DEP, the antecedent $A$ logically relates to the presupposition $p$ (here, being a carpenter relates to using beading tools). In IND, $A$ and $p$ are unrelated (finishing a meeting has no connection to driving sports cars). Presupposition triggers are underlined.}
\end{table}

We expanded the dataset to approximately 8,500 sentence pairs by modifying the original 900 examples. The expansions focused on both structural and semantic properties of the conditionals, including syntactic variation, changes in trigger–hypothesis relatedness, and premise-level contextual modifications. The original labels were largely retained as E, with a small subset reclassified as N. In addition, we added a new column containing theory-based labels, which are consistently marked as neutral for all sentence pairs. This is based on the theoretical accounts of the proviso problem that accept only the conditional form of presupposition, in contrast to the unconditional hypotheses represented in our dataset. 

The entailment labels used in our dataset follow the annotation scheme of the CONFER dataset \citeplanguageresource{CONFER}, on which our dataset is built. For the original 900 sentence pairs, we retain the labels provided in CONFER. During dataset expansion, modified examples were created and reviewed by a computational linguist to ensure that the semantic relation between the premise and the hypothesis remained consistent with the original annotation scheme. Because the modifications were designed to preserve the original entailment relation, the labels were largely retained as \textit{E}, with a small subset reclassified as \textit{N} when the modification affected the inference relation. Since the annotation and verification were conducted by a single annotator, inter-annotator agreement does not apply. 

The additional theory-based label column was generated deterministically based on theoretical accounts of the proviso problem \citep{Singh2007, vonFintel-2008}. Under these accounts, presuppositions project conditionally from conditional antecedents, which means that unconditional hypotheses of the form used in our dataset are predicted to be neutral with respect to the premise. Consequently, all sentence pairs receive a neutral theory-based label. The resulting dataset is organized into four subsets.\footnote{The dataset and experimental results are available at \url{https://github.com/Conditional-NLI/PROVISER}.} Table \ref{tab:dataset-examples} provides examples from each subset. In this section, we describe the structure of each subset in more detail.

\textbf{Naming conventions.} Throughout this paper, we use the following convention to refer to specific data types. DEP conditionals include only possessive triggers (e.g., \textit{his wife}), whereas IND conditionals include either possessive triggers or the adverbial trigger \textit{again}. Accordingly, each IND example carries either the \texttt{-poss} (for possessive) or \texttt{-again} suffix. We further distinguish between the original examples drawn directly from CONFER (e.g., \textit{DEP}, \textit{IND-poss}, \textit{IND-again}) and their modified variants created for our experiments (e.g., \textit{DEP-mod}, \textit{IND-poss-mod}, \textit{IND-again-mod}). When no \texttt{-mod} suffix is present, the example refers to the original, unmodified instance from CONFER. 

\subsection{Subset 1: Original Sentences}
This subset used the initial 900 NLI pairs without modifications as a baseline to evaluate models' performance.

\subsection{Subset 2: Structural Variation}
\label{subset2}

To test how structural manipulation affects presupposition projection, we created three modified versions of each original sentence: (1) Conjunction: adding a conjunct to the antecedent (e.g., \textit{If $A$ and $B$, then $C$}); (2) Disjunction: rephrasing the conditional as \textit{Either not-$A$ or $B$}; and (3) Belief Embedding: embedding the consequent under an attitude verb (e.g., \textit{X believes that $B$}).

These modifications are based on established findings in presupposition theory showing that presuppositions behave differently across various sentence structures \citep{Karttunen-1973, Heim-1983}. In conjunctions, presuppositions from both parts typically survive. In disjunctions, presuppositions may be blocked or filtered out. Finally, under belief verbs, presuppositions may or may not project depending on the specific verb used.

\subsection{Subset 3: Trigger–Hypothesis Relatedness}
\label{subset 3}

In this subset, we modified the trigger phrase by manipulating its semantic similarity to the hypothesis. The original examples naturally contained triggers that were semantically \textit{related} to their hypotheses. Using WordNet \citep{WordNet} and ConceptNet \citeplanguageresource{ConceptNet} relations (e.g., \texttt{is-a}, \texttt{part-of}, \texttt{hyponymy}), we generated two additional variants, \textit{somewhat related} and \textit{unrelated}, by substituting key lexical items in the presupposition trigger, thus creating three levels of semantic relatedness for each example.

For example, consider the original premise \textit{If Matt is a scuba diver, he'll bring his wetsuit} with the hypothesis \textit{Matt has a wetsuit}. Here are examples with two degrees of relatedness between the hypothesis and the trigger phrase:

\begin{itemize}
    \item \textbf{Somewhat related}: If Matt is a scuba diver, he'll bring his garment. (garment is a hypernym of wetsuit)
    \item \textbf{Unrelated}: If Matt is a scuba diver, John's friend will bring her wetsuit. (mismatched agent and possessor)
    
\end{itemize}

{\small \textit{*Hypothesis for both: Matt has a wetsuit.}}

\smallskip

As the semantic relationship between the trigger and hypothesis weakens, the gold labels were changed to show the reduced support for entailment. In this example, both the \textit{somewhat related} and \textit{unrelated} variants received N labels instead of the original E, since the premise no longer provides clear evidence for the hypothesis.

\subsection{Subset 4: Context–Trigger Relatedness}
\label{subset 4}
This subset examines how the semantic relationship between different parts of the conditional affects presupposition projection. We modified the logical connection between the antecedent and consequent to be either logically related or unrelated.\footnote{The logical relation between $A$ and $p$ remained the same, therefore we still have the same DEP and IND sentence types.}

Each sentence contains a presupposition trigger embedded within a conditional structure. We created the two variants by altering the semantic relationship between the conditional's antecedent and consequent. This approach is motivated by research showing that presuppositions can be sensitive to the broader discourse context and the logical coherence of the surrounding material \citep{geurts1999presuppositions, beaver2001presupposition}.

The following examples illustrate these two variants:

\begin{itemize}
    \item \textbf{Related:} If Sarah attends a movie festival, she'll never watch Star Wars again.
    
    \item \textbf{Unrelated:} If Sarah attends the conference, she'll never watch Star Wars again.
\end{itemize}

{\small \textit{*Hypothesis for both: Sarah has watched Star Wars before.}}

\smallskip

To facilitate attention analysis (detailed in Section \ref{evaluation_metrics}), we annotated two key regions in each example: K1, the noun phrase preceding the presupposition trigger in the hypothesis (e.g., \textit{Star Wars} in the examples above), and K2, the contextual phrase in the consequent that provides semantic context (e.g., \textit{attends a movie festival}). We use these marked regions to analyze how models link the trigger to its surrounding context.

According to presupposition theory \citep{Karttunen-1973, Heim-1983}, the presupposed content should project regardless of the logical relationship between the antecedent and consequent, since presuppositions are typically preserved across different contexts. However, we hypothesize that language models may be sensitive to these contextual manipulations, potentially showing different labeling patterns when the conditional's parts are semantically unrelated, even though the presupposition itself remains constant.

\begin{table*}[ht]
\centering
\scriptsize  
\renewcommand{\arraystretch}{0.85}  
\begin{tabular}{@{}p{0.9cm}p{7cm}p{3cm}p{0.9cm}p{0.9cm}@{}}
\toprule
\textbf{Subset} & \textbf{Premise} & \textbf{Hypothesis} & \textbf{Human} & \textbf{Theory} \\
\midrule
1 & 
\textbf{DEP:} If John is a scuba diver, he'll bring his wetsuit. 
\smallskip

\textbf{IND:} If Lisa finishes meeting early, she'll never drive a sports car again. &
John has a wetsuit. 
\smallskip

Lisa has driven a sports car before. &
E & N \\
\midrule
2 & 

\textbf{Conjunction:} If Sarah attends the conference and her friend meets her at the hotel, it's okay. 
\smallskip

\textbf{Disjunction:} Either Sarah doesn't attend conference, or she and her friend do. 
\smallskip

\textbf{Belief:} Sarah believes her friend is upstairs. &
Sarah has a friend. &
E & N \\
\midrule
3 &
\textbf{Related:} If Nadia attends conference, she'll never watch Star Wars again. 
\smallskip

\textbf{Somewhat related:} If Nadia attends conference, she'll never watch a George Lucas movie again. 
\smallskip

\textbf{Unrelated:} If Nadia attends conference, she'll never watch Titanic again. &
Nadia has watched Star Wars. &
E/N$^{*}$ & N\\
\midrule
4 &
\textbf{Related:} If Ali attends conference, his brother will meet her at university. 
\smallskip

\textbf{Unrelated:} If Ali attends conference, his brother will get rid of his old car. &
Ali has a brother. &
E & N \\
\bottomrule
\end{tabular}
\caption{Examples from the four subsets with corresponding human and theory-based labels. $^{*}$In Subset 3, the human label is E for the related variant where the trigger matches the hypothesis, and N for somewhat related and unrelated variants where the modified trigger no longer semantically supports the hypothesis.}
\label{tab:dataset-examples}
\end{table*}

\section{Experiments}
\label{sec:4}

We evaluated four language models to determine whether they process presuppositions in conditionals according to human judgments or theory-based predictions. Our evaluation combines classification accuracy with explainability analyses across controlled varied datasets. Human labels reflect empirical judgments from the original examples, while theory-based labels follow the conditional projection patterns predicted by formal semantic theories discussed in Section~\ref{sec:introduction}. Below, we describe the models, evaluation metrics, and experimental procedures.

\subsection{Models and Setup}

Our experiments use RoBERTa-large-MNLI, DeBERTa-large-MNLI, Llama-3.2-1B, and Gemma-3-1B. RoBERTa and DeBERTa were selected as they are NLI-specific models, pre-trained on MultiNLI\citeplanguageresource{MultiNLI}.Llama and Gemma were included as they are recent, open-weight large-language models. All models were fine-tuned on the CONFER training set for one epoch with a learning rate of 5e-5.

\subsection{Evaluation Metrics}
\label{evaluation_metrics}

We use both classification accuracy and explainability metrics to assess whether model predictions align with human or theory-based labels and to determine which linguistic features drive those predictions.

\textbf{Classification Accuracy} is calculated against both human labels and theory-based labels.

\smallskip

\textbf{Integrated Gradients (IG)} \citep{Sundararajan-2017} measures token-level attribution by integrating the gradients of the model output along a path from a baseline input $x'$ to the actual input $x$:
\begin{equation}
\small
\text{IG}_i(x) = (x_i - x'_i) \int_{0}^{1} \frac{\partial F(x' + \alpha (x - x'))}{\partial x_i} \, d\alpha
\label{eq:ig}
\end{equation}
\noindent where $x$ is the input, $x'$ the baseline, $x_i$ a token, $\alpha \in [0,1]$ the interpolation factor, and $\tfrac{\partial F}{\partial x_i}$ the partial derivative of the model output with respect to token $i$.

IG applies the same backpropagation mechanism that the model uses during optimization to compute gradients of the output with respect to the input. Like perturbation-based methods such as LIME \citep{LIME} and SHAP \citep{SHAP}, it identifies important input features, but does so more cost-effectively by using gradient information rather than iterative input modifications \citep{bastings-filippova-2020-elephant}.

\begin{figure*}[t]
  \centering
  \fbox{%
    \includegraphics[width=0.92\textwidth]{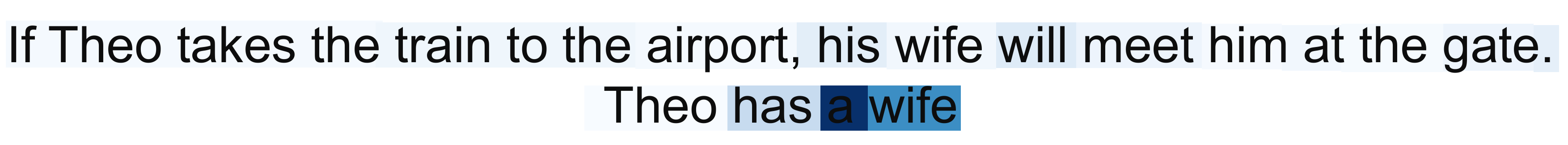}
  }
  \caption{IG visualization showing token-level attribution. Darker shading indicates higher IG values, with the presupposition trigger receiving the strongest attribution.}
  \label{fig:IG-example}
\end{figure*}

Using this method, we can address an important question: do models focus on the same linguistic elements that humans consider crucial for presuppositional reasoning? In presuppositional entailment, trigger phrases provide the critical link between conditionals and their presuppositions. For example, in \textit{If Randolf is a carpenter, he'll use his beading tools for designing}, the phrase \textit{his beading tools} triggers the inference that Randolf has beading tools. Without this trigger, the entailment would not arise. Figure~\ref{fig:IG-example} illustrates how IG highlights such tokens.

From the IG scores, we derive another additional metric called the trigger IG ratio, which quantifies the proportional influence assigned to the trigger words relative to the average influence of all tokens.

\smallskip

\textbf{Context–Trigger Attention Metrics} complement IG by showing how models encode relationships between phrases rather than individual token salience. While IG shows which tokens matter for predictions, attention patterns indicate whether models connect presupposition triggers with their contextual dependencies, which is essential for pragmatic reasoning in conditionals.

For Subset 4, we compute two metrics:
\begin{itemize}
    \item \textbf{K1$\rightarrow$K2 attention:} measures attention flow from the noun phrase preceding the presupposition trigger in the hypothesis (K1) to the contextual phrase (K2), normalized by overall attention. This assesses whether models establish stronger connections between semantically related elements.
    \item \textbf{K2$\rightarrow$special token attention:} measures attention from K2 to delimiter tokens ([SEP], \textless bos\textgreater, [CLS]), normalized by average special token attention. This indicates whether models treat context as structurally significant for sentence representation.
\end{itemize}

\smallskip

\textbf{T-test for Significance} assesses whether modifications significantly affect trigger IG ratios. We perform paired t-tests (reported with p-values) comparing original and modified examples ($\alpha$ = 0.05), indicating which structural and semantic variations reliably affect model attention patterns. FDR multiple-comparison correction was applied.

\subsection{Zero-Shot Evaluation}

To assess whether models can handle presuppositional reasoning based on their pretraining without task-specific fine-tuning, we evaluated all four models on subsets 1-4 in a zero-shot setting. RoBERTa and DeBERTa, pre-trained on MultiNLI, were tested directly as a three-way classification task (Entailment, Neutral, Contradiction). Gemma and LLaMA were evaluated through zero-shot prompting using the instruction shown in Figure~\ref{fig:prompt}. We computed classification accuracy against both human and theory-based labels to establish baseline performance before fine-tuning.

\subsection{Structural Variations}

To examine whether structural variations affect model sensitivity to presupposition triggers, we fine-tuned all four models on CONFER and evaluated them on Subset 2, which contains three structural modifications explained in \ref{subset2}. 
We computed classification accuracy alongside the trigger IG ratio to assess whether these structural changes shift the importance models place on presuppositional content.

\setlength{\fboxsep}{3pt}   
\setlength{\fboxrule}{0.0pt} 

\begin{figure}[t]
\centering
\begin{promptbox}
\begin{Verbatim}
Consider the truth value of the premise. 
If the premise is true, does it necessarily 
mean that the hypothesis must also be true?

Output E if the hypothesis must always be true.
Output C if the hypothesis must always be false.
Output N if the hypothesis may be either 
true or false.

Do not output anything other than E, C, or N.

Premise: {premise}
Hypothesis: {hypothesis}
Output:
\end{Verbatim}
\end{promptbox}
\caption{Three-way classification prompt used for zero-shot evaluation of Gemma and LLaMA models.}
\label{fig:prompt}
\end{figure}

\subsection{Lexical-Semantic Substitutions}

Experiments on Subset 3 evaluate whether models process presupposition triggers based on semantic content or surface-level position. As described in Section~\ref{subset 3}, we created three relatedness levels by substituting trigger phrases with semantically related, somewhat related, or unrelated alternatives, updating gold labels from Entailment to Neutral when the semantic relationship no longer holds. We measured classification accuracy and trigger IG ratio scores to assess whether models appropriately adjust their predictions when trigger semantics change, or continue to rely on positional heuristics.

\subsection{Contextual Modifications}

To assess whether models are sensitive to the logical relationship between conditional components, we evaluated fine-tuned models on Subset 4, which manipulates premise-level context while keeping the presupposition trigger constant. As described in Section~\ref{subset 4}, we created variants where the antecedent and consequent events are either logically related or unrelated. We measured classification accuracy alongside K1$\rightarrow$K2 attention and K2$\rightarrow$special token attention to determine whether models maintain appropriate focus on presuppositional triggers regardless of contextual distractors, or whether unrelated context phrases interfere with presupposition processing.

\section{Results and Analysis}
\label{sec:5}

\subsection{Models Align with Human Judgments and Not Theories}

\begin{figure*}[t]
  \centering
  \includegraphics[width=\textwidth]{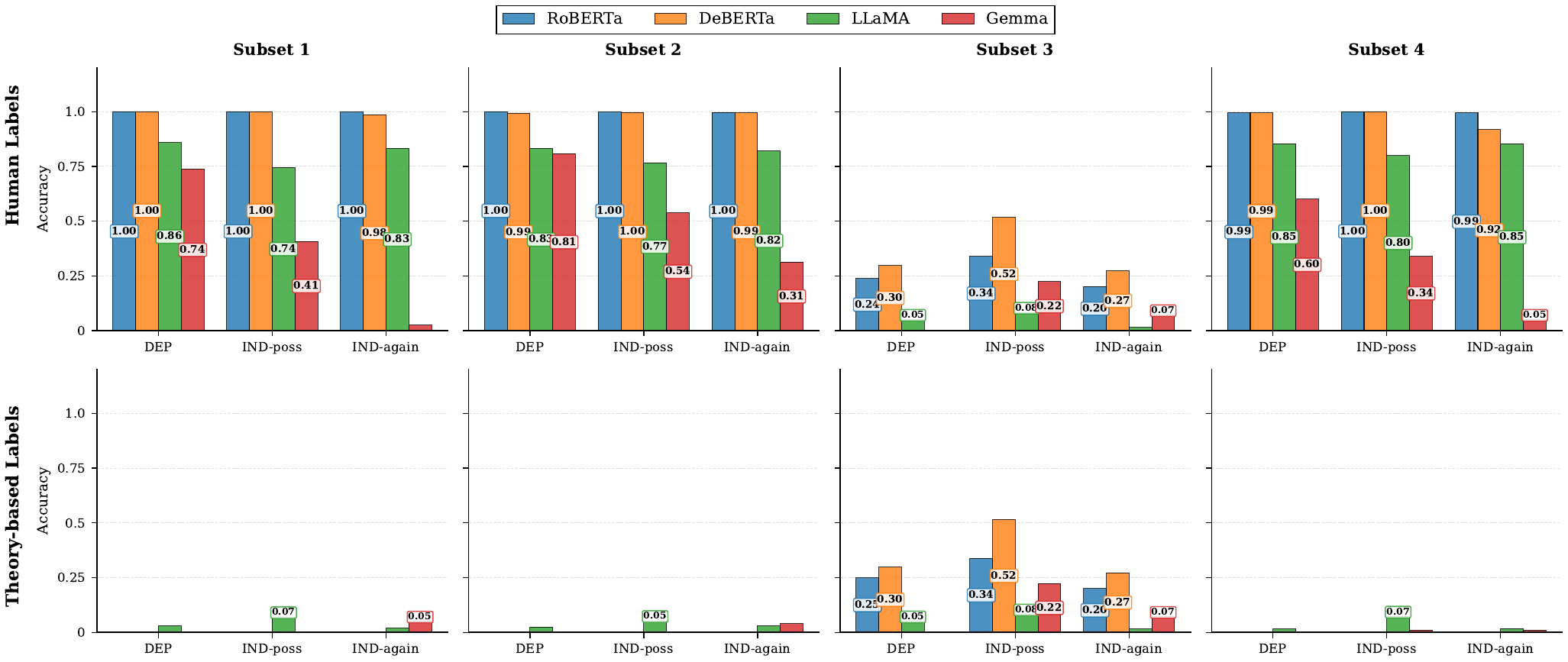}
  \caption{Models’ accuracies in zero-shot evaluation across four subsets, using human and theory-based labels.}
  \label{fig:model-accuracy}

\end{figure*}

In zero-shot evaluation, we assessed whether pre-trained models exhibit human-like or theory-aligned presuppositional reasoning. As indicated in Figure \ref{fig:model-accuracy}, RoBERTa and DeBERTa achieved perfect accuracy against human labels on original examples (Subset 1). In contrast, they showed 0\% accuracy against theory-based labels. The results indicate complete alignment with human unconditional interpretations rather than the conditional presuppositions predicted by formal semantics.In this experiment, LLaMA and Gemma underperformed compared to RoBERTa and DeBERTa. LLaMA achieved 74-86\% accuracy against human labels, while Gemma struggled across all sentence types in this subset.

On modified examples (Subsets 2-4), models maintained alignment with human judgments for Subsets 2 (structural variations) and 4 (contextual modifications), with accuracy remaining above 90\% for RoBERTa and DeBERTa. However, performance dropped dramatically on Subset 3 (semantic substitutions), where models correctly identified only 24-52\% of examples as N despite the broken semantic relationship between modified triggers and hypotheses. This shows that models rely on structural and positional cues rather than semantic content when processing presupposition triggers.

\subsection{Structural Modifications Maintain Trigger Focus Despite Complexity}

\begin{figure}[t]
    \centering
    \includegraphics[width=\columnwidth]{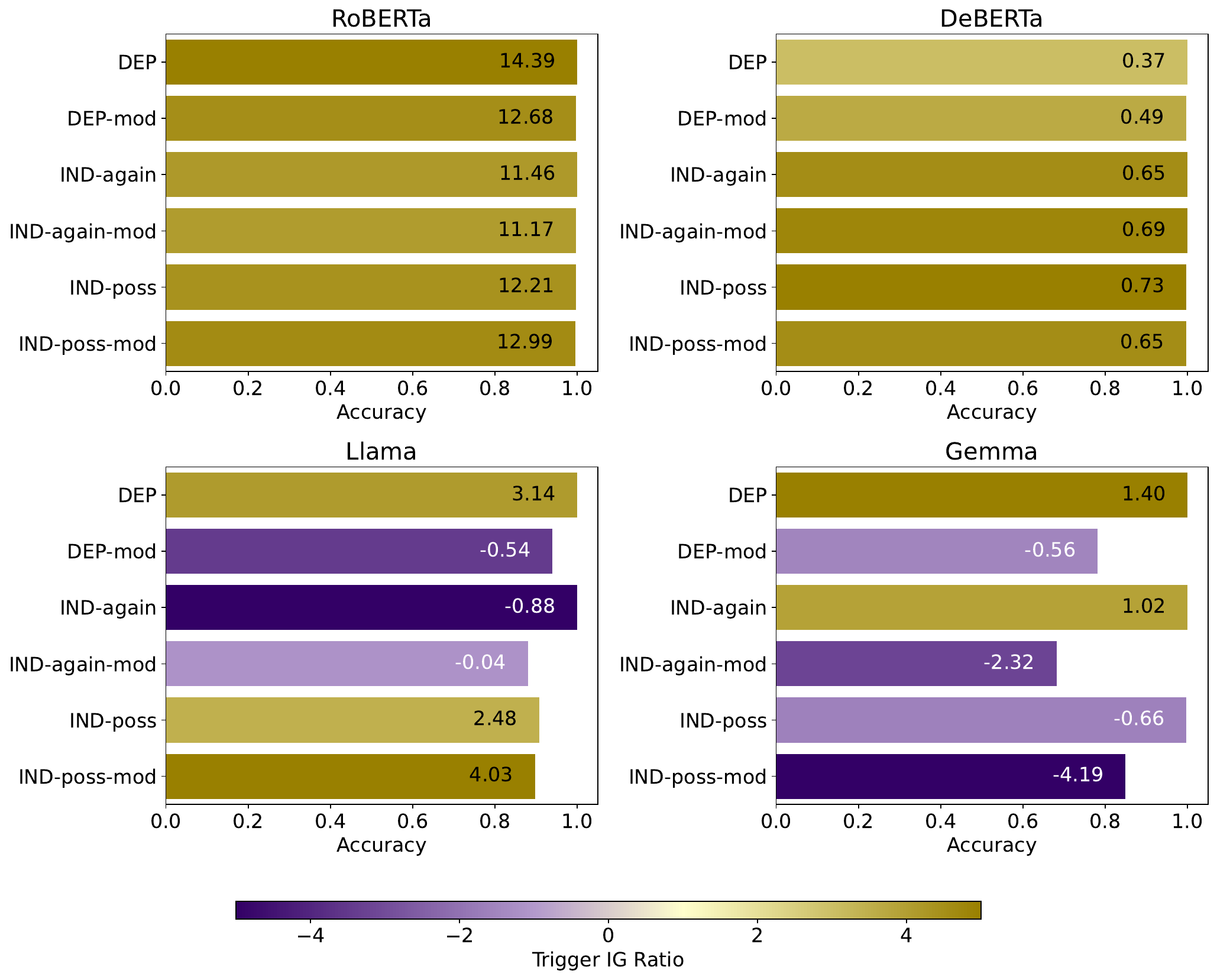}
    \caption{Accuracy across sentence types in Subset 2. The horizontal axis shows accuracy percentages, and the values inside the bars indicate the corresponding trigger IG ratios.}
    \label{fig:part4A-subset2}
\end{figure}

Fine-tuned models maintained high classification accuracy on Subset 2, with RoBERTa and DeBERTa achieving 99-100\% accuracy on both original and modified examples. However, models showed different attention patterns. RoBERTa indicated the highest trigger IG ratios across both original and structurally embedded examples, indicating strong focus on presupposition triggers regardless of syntactic complexity (Figure~\ref{fig:part4A-subset2}). Statistical tests showed no significant differences in trigger IG ratios between original and modified examples (p > 0.05), suggesting RoBERTa's trigger sensitivity is robust to structural variations.

DeBERTa showed lower trigger IG ratios (<1) across all examples, indicating below-average attention to triggers, though accuracy remained nearly perfect. LLaMA and Gemma showed greater variability in trigger IG, including occasional negative values, with modest accuracy drops.

\subsection{Models Rely on Trigger Position Over Semantic Content}

Models achieved high accuracy on Subset 3's unmodified original examples but failed dramatically when trigger phrases were semantically modified. Despite human labels changing to N, models continued predicting E, showing reliance on structural position over semantic content.

Performance varied by trigger type and model. IND-again examples proved most difficult, with all models achieving below 4\% accuracy, nearly complete misclassification. On DEP-poss examples, models performed marginally better with RoBERTa reaching 22\% accuracy while others remained below 15\%. IND-poss examples were relatively easier: RoBERTa and Gemma achieved above 35\% accuracy, and LLaMA surprisingly exceeded 80\%. However, DeBERTa misclassified all IND-poss examples.

\begin{figure}[t]
    \centering
    \includegraphics[width=\columnwidth]{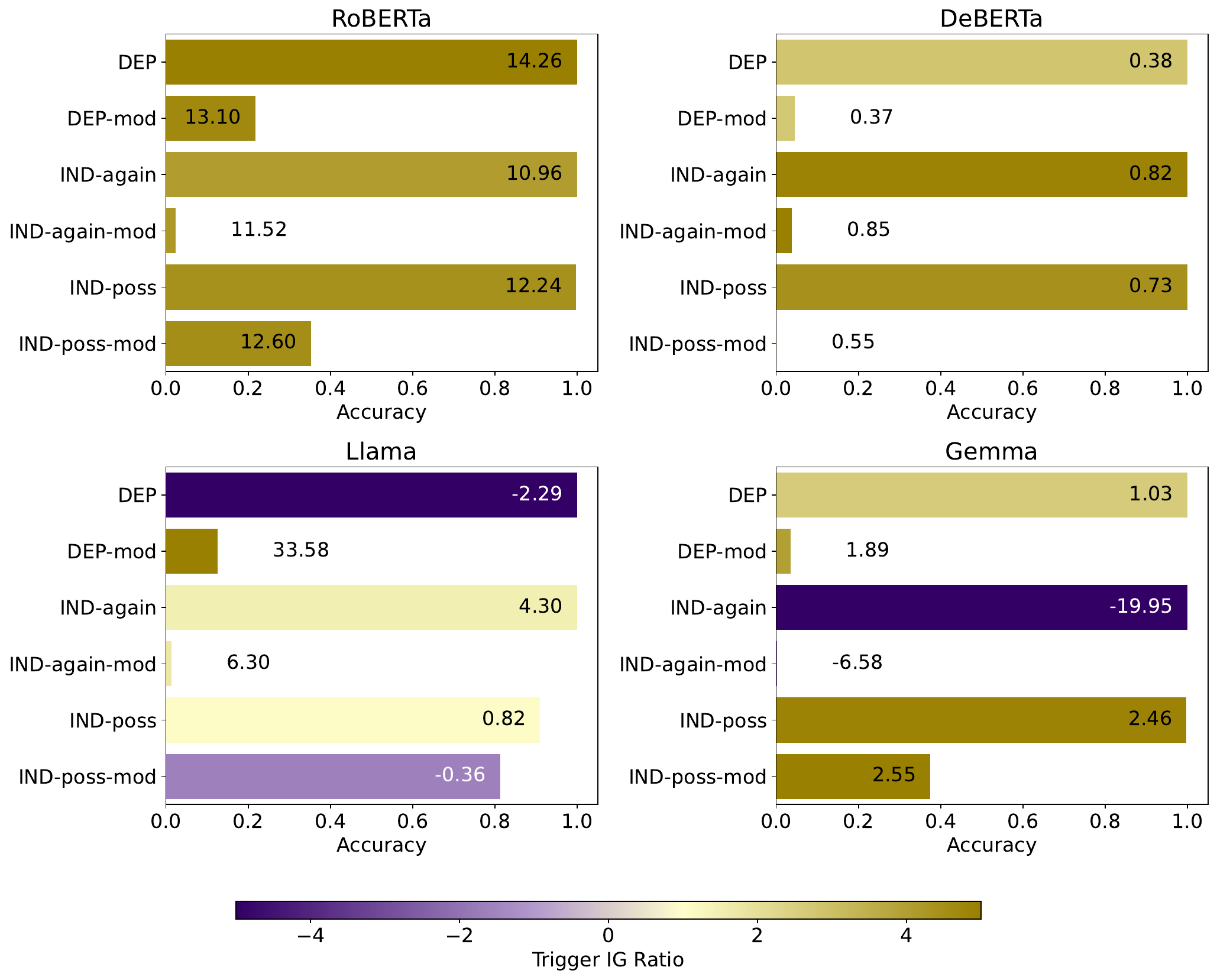}
    \caption{Accuracy and trigger IG ratios across sentence types in Subset 3}
    \label{fig:part4B-subset3}
\end{figure}

Trigger IG analysis confirmed models' reliance on structural position (Figure~\ref{fig:part4B-subset3}). Across all forms, significant differences did not emerge between the way models treat trigger phrases in the original examples compared to the modified examples (p > 0.05), despite semantic changes. RoBERTa maintained high IG ratios regardless of semantic fit, while DeBERTa showed consistently low values, with ratios stable between 0.8-0.85 even in IND-again examples. LLaMA and Gemma exhibited variability, with LLaMA occasionally increasing IG in unrelated contexts and Gemma showing negative values. Moreover, LlaMa achieving 80\% accuracy on IND-poss examples, does see a slight negative IG shift, correctly indicating that the trigger phrase no longer carries importance in the neutral relationships. The overall stable attribution despite semantic manipulation confirms that models rely on trigger position rather than meaning.

\begin{figure}[t]
    \centering
    \includegraphics[width=\columnwidth]{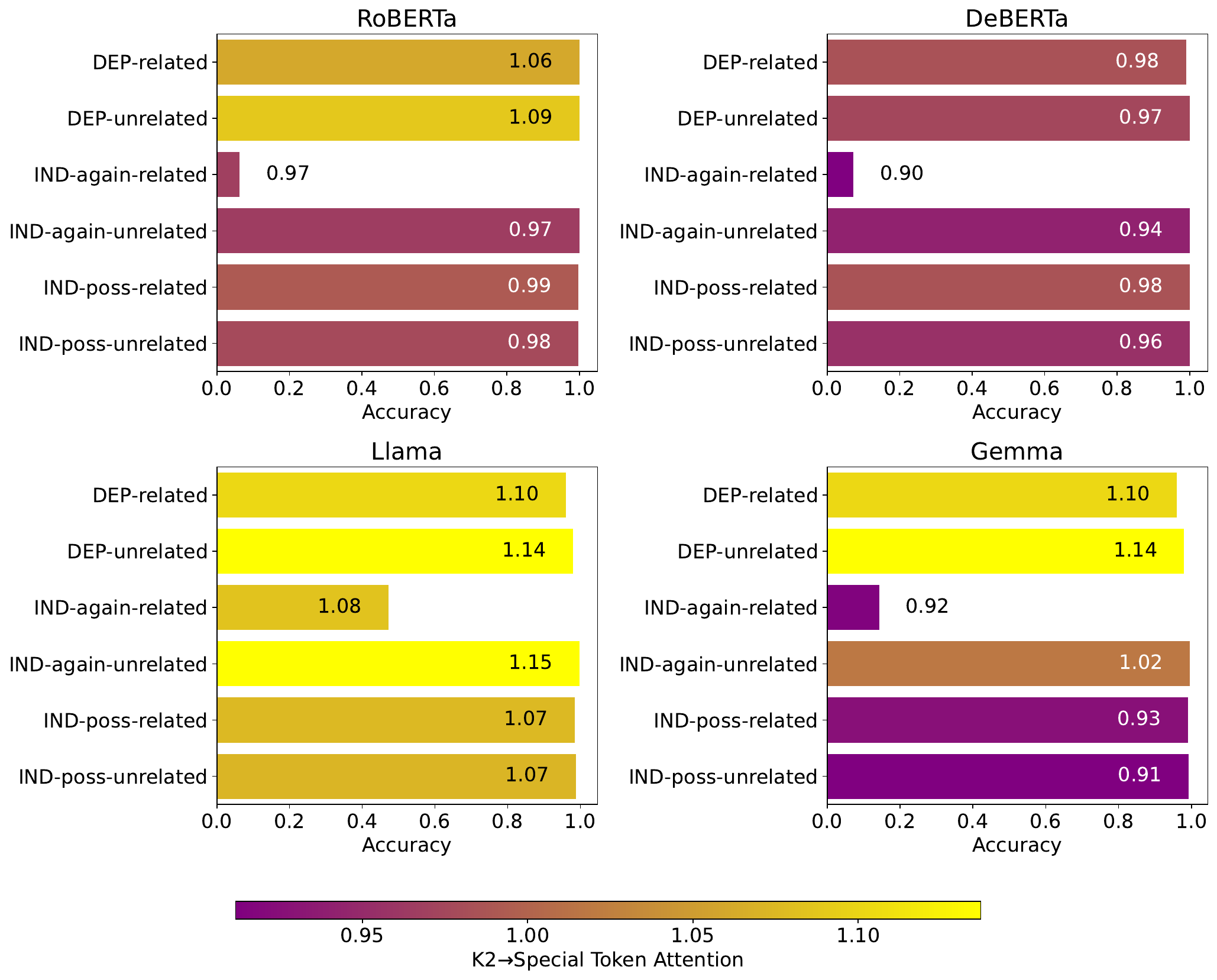}
    \caption{Subset 4 accuracy by sentence type and model, with bar colours indicating K2->special token attention on the corresponding examples.}
    \label{fig:k1k2}
\end{figure}

\subsection{Contextual Modifications Expose Training Set Overfitting}

Models maintained perfect accuracy (\textasciitilde100\%) on modified DEP and IND-poss examples in Subset 4. They correctly predicted E regardless of whether contextual phrases (K2) were semantically related or unrelated to the noun phrase in the hypothesis (K1). However, accuracy decreased significantly on modified IND-again examples, with LLaMA achieving only 47\% and the other three models below 15\% (Figure~\ref{fig:k1k2}). Models incorrectly predicted N instead of E despite only irrelevant context words changing. Since zero-shot models achieved >80\% accuracy on these same examples (except Gemma), this failure appears to result from CONFER fine-tuning. CONFER includes examples where semantically related antecedents pair with \textit{again} triggers but carry N labels due to superset relationships in the antecedent. Models likely learned the spurious pattern "related antecedent + again = N" from these examples, causing confusion with the structurally similar IND-again modifications.

Attention analysis indicates distinct patterns by trigger type. K1$\rightarrow$K2 attention showed significant variation across all trigger types. For DEP and IND-poss examples, these attention shifts corresponded with models' accuracy patterns. In IND-again examples, models placed significantly less attention from the trigger noun phrase to context in modified variants compared to original ones, showing the substantial accuracy drops observed in this condition.

\begin{figure}[t]
    \centering
    \includegraphics[width=0.5\textwidth]{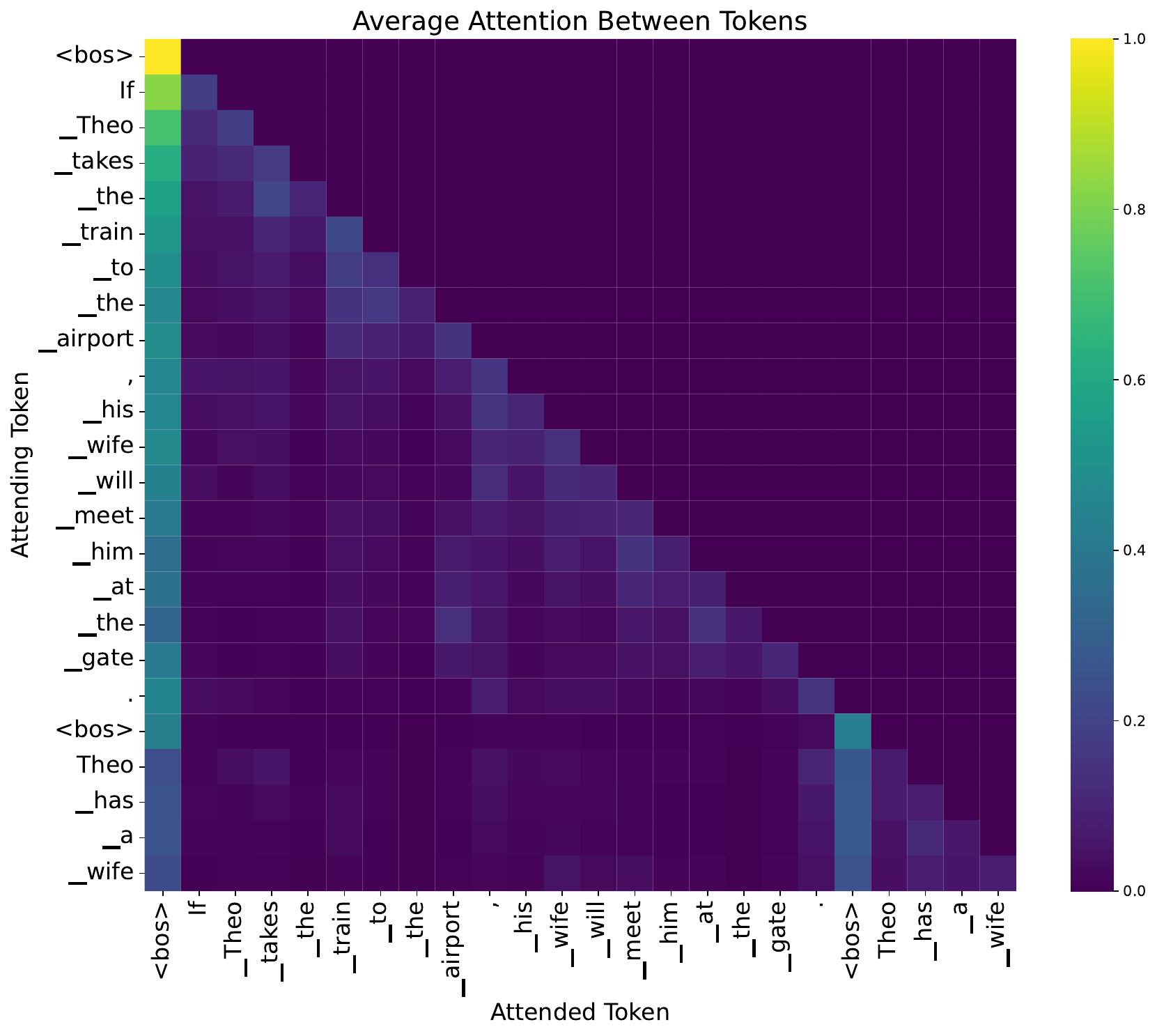}
    \caption{Average attention between tokens across all layers and heads in Gemma, for the sentence \textit{If Theo's train is delayed, his wife will wait for him at the station}, and its hypothesis \textit{Theo has a wife}. Brighter cells indicate higher attention values. First column corresponds to the special token attention.}
    \label{fig:attention-example}
\end{figure}

Models also significantly increased K2→special token attention in modified IND-again examples compared to originals (p < 0.08 for RoBERTa, p < 0.001 for others), indicating these contextual modifications affected sentence-level representations. While RoBERTa and Gemma showed some attention shifts in other subsets, only IND-again modifications produced consistent, significant changes across all four models, the same subset where accuracy decreased. This correlation between shifted attention patterns and misclassification shows overfitting to training patterns where "related context + again" was associated with N labels, leading models to apply this association inappropriately. Figure \ref{fig:attention-example} illustrates the average attention between tokens in Gemma.

\section{Discussion and Conclusion}
\label{sec:discussion}

We introduced a diagnostic resource for evaluating how language models process presuppositional reasoning in conditionals, addressing the proviso problem. We created a dataset of 8,500 examples with controlled linguistic variations and an evaluation framework combining classification accuracy with explainability methods. This framework, facilitates assessment of whether models process presuppositions through semantic understanding or shallow heuristics, with comparisons to human reasoning patterns.

Our evaluation shows that while models align with human judgments, they achieve this through pattern matching rather than semantic reasoning. Diagnostic tests revealed three main limitations. First, models dramatically fail when trigger phrases are replaced with semantically unrelated alternatives. Second, accuracy alone can be misleading for assessing model reasoning. On Subset 2, RoBERTa achieves 99\% accuracy by focusing more on trigger words (IG scores 11-17), while DeBERTa achieves identical accuracy while putting minimal importance on the trigger (IG <1). However, Subset 3 then reveals the extent to which DeBERTa is fallible under the modifications, achieving an accuracy of 0\%, while RoBERTa is more robust, still capable of achieving an accuracy between 20\% and 50\% on Subset 3. Third, fine-tuning can introduce misleading correlations that degrade performance on structurally similar (but logically different) examples.

This work demonstrates that evaluating pragmatic reasoning in language models requires diagnostic resources paired with multi-method assessment frameworks. While accuracy metrics suggest strong performance, our explainability analysis shows models rely on structural patterns rather than semantic understanding. Future work can extend this diagnostic approach to other pragmatic phenomena such as scalar and conventional implicature. Moreover, exploring task formats beyond NLI would assess whether findings generalize across evaluation paradigms. Additionally, parallel psycholinguistic experiments could test longstanding semantic theories empirically by directly comparing human processing patterns with model behavior.

\section{Limitations}

This study focuses on a specific type of conditional sentences structured in the form \textit{If $A$, $B$\textsubscript{$p$}}, where the presupposition trigger appears in the consequent. The conditionals include only two trigger types: possessive pronouns and the adverbial trigger \textit{again}. These triggers were selected because they occur frequently in the source CONFER dataset~\cite{azin2025confer} and yield presuppositions that can be directly restated as standalone hypotheses, making them suitable for the NLI formulation used in our experiments. This design choice means the dataset does not capture the full range of presuppositional phenomena in natural language, including other trigger types (e.g., factive verbs, aspectual verbs, clefts), different conditional constructions, or interactions with broader discourse context.

Our evaluation is limited to four language models, and findings may not generalize to other model architectures, sizes, or training paradigms. The NLI task format, while appropriate for testing inferential relationships, represents only one way of probing presuppositional reasoning and may not fully reflect how models process presuppositions in generation or other downstream tasks.

\section{Ethical Considerations}

All experimental materials in this study consist of everyday scenarios with no sensitive content, stereotypes, or references to real individuals. The sentences were designed only to test linguistic phenomena. As this is a diagnostic evaluation study rather than a deployment of models for real-world applications, the research poses minimal ethical risks. The dataset and code will be made publicly available upon publication for transparency and reproducibility.

\newpage
 
\nocite{*}
\section{References}\label{sec:reference}

\bibliographystyle{lrec2026-natbib}
\bibliography{references}

\end{document}